\documentclass{scrartcl}

\PassOptionsToPackage{numbers, compress}{natbib}
\usepackage{hyperref}
\usepackage{url}
\usepackage[utf8]{inputenc} 
\usepackage[T1]{fontenc}    
\usepackage{booktabs}       
\usepackage{amsfonts}       
\usepackage{nicefrac}       
\usepackage{microtype}      
\usepackage{graphicx}
\usepackage{algorithm}
\usepackage[noend]{algpseudocode}
\usepackage{amsmath}
\usepackage{multirow}
\usepackage{tabularx}
\usepackage{wrapfig}
\usepackage{lipsum}
\usepackage{array}
\usepackage{subcaption}
\usepackage{mathtools}
\usepackage{icomma}
\usepackage{siunitx}
\usepackage{csquotes}
\usepackage{amsmath,amsthm,amssymb}

\usepackage{pst-node}
\usepackage{tikz-cd}

\newcommand\cut[1]{}

\newcolumntype{C}[1]{>{\centering\arraybackslash}m{#1}}
\newcolumntype{R}[1]{>{\raggedleft\arraybackslash}m{#1}}

\newcommand{\be}{\begin{equation}}
\newcommand{\ee}{\end{equation}}
\newcommand{\bea}{\begin{eqnarray}}
\newcommand{\eea}{\end{eqnarray}}
\newcommand{\beaa}{\begin{eqnarray*}}
\newcommand{\eeaa}{\end{eqnarray*}}

\DeclareMathAlphabet{\mathpzc}{OT1}{pzc}{m}{n}

\graphicspath{{figures/}}

\DeclarePairedDelimiterX{\divx}[2]{\big(}{\big)}{%
  #1\;\delimsize\|\;#2%
}

\usepackage{xspace}

\newcommand{\betavae}{\ensuremath{\beta}-VAE\xspace}

\newboolean{draftmode}
\setboolean{draftmode}{true}

\usepackage{color}
\newcommand{\mycomment}[3]{{\textcolor{#3}{[#1 #2]}}}
\newcommand{\ihmarker}{{\textcolor{blue}{\ensuremath{^{\textsc{I}}_{\textsc{H}}}}}}
\newcommand{\almarker}{{\textcolor{red}{\ensuremath{^{\textsc{A}}_{\textsc{L}}}}}}
\newcommand{\damarker}{{\textcolor{magenta}{\ensuremath{^{\textsc{D}}_{\textsc{A}}}}}}
\newcommand{\dpfmarker}{{\textcolor{brown}{\ensuremath{^{\textsc{D}}_{\textsc{P}}}}}}
\newcommand{\apmarker}{{\textcolor{cyan}{\ensuremath{^{\textsc{A}}_{\textsc{P}}}}}}
\newcommand{\srmarker}{{\textcolor{orange}{\ensuremath{^{\textsc{S}}_{\textsc{R}}}}}}
\newcommand{\lmmarker}{{\textcolor{purple}{\ensuremath{^{\textsc{L}}_{\textsc{M}}}}}}
\newcommand{\nwmarker}{{\textcolor{olive}{\ensuremath{^{\textsc{N}}_{\textsc{W}}}}}}
\newcommand{\cbmarker}{{\textcolor{teal}{\ensuremath{^{\textsc{C}}_{\textsc{B}}}}}}

\ifthenelse{\boolean{draftmode}}{
 \newcommand{\ih}[1]{\mycomment{\ihmarker}{#1}{blue}}
 \newcommand{\al}[1]{\mycomment{\almarker}{#1}{red}}
 \newcommand{\da}[1]{\mycomment{\damarker}{#1}{magenta}}
 \newcommand{\dpf}[1]{\mycomment{\dpfmarker}{#1}{brown}} 
 \newcommand{\ap}[1]{\mycomment{\apmarker}{#1}{cyan}} 
 \newcommand{\sr}[1]{\mycomment{\srmarker}{#1}{orange}} 
 \newcommand{\lm}[1]{\mycomment{\lmmarker}{#1}{purple}} 
 \newcommand{\nw}[1]{\mycomment{\nwmarker}{#1}{olive}} 
 \newcommand{\cb}[1]{\mycomment{\cbmarker}{#1}{teal}} 
}{
 \newcommand{\ih}[1]{}
 \newcommand{\al}[1]{}
 \newcommand{\da}[1]{}
 \newcommand{\dpf}[1]{} 
 \newcommand{\ap}[1]{}
 \newcommand{\sr}[1]{}
 \newcommand{\lm}[1]{}
 \newcommand{\nw}[1]{}
 \newcommand{\cb}[1]{}
}

\makeatletter
\def\BState{\State\hskip-\ALG@thistlm}
\makeatother

\title{Towards a Definition of \\ Disentangled Representations}

\author{
 Irina~Higgins$^*$, David~Amos$^*$, David~Pfau, Sebastien~Racaniere,  \\ Loic~Matthey, Danilo~Rezende, Alexander~Lerchner\\
 DeepMind\\
 \texttt{\{irinah,davidamos,pfau,sracaniere,}\\
 \texttt{lmatthey,danilor,lerchner\}@google.com}
}

\begin{document}
\maketitle

\begin{abstract}
How can intelligent agents solve a diverse set of tasks in a data-efficient manner? The disentangled representation learning approach posits that such an agent would benefit from separating out (disentangling) the underlying structure of the world into disjoint parts of its representation. However, there is no generally agreed-upon definition of disentangling, not least because it is unclear how to formalise the notion of world structure beyond toy datasets with a known ground truth generative process. Here we propose that a principled solution to characterising disentangled representations can be found by focusing on the \emph{transformation} properties of the world. In particular, we suggest that those transformations that change only some properties of the underlying world state, while leaving all other properties invariant, are what gives exploitable structure to any kind of data. Similar ideas have already been successfully applied in physics, where the study of symmetry transformations has revolutionised the understanding of the world structure. By connecting symmetry transformations to vector representations using the formalism of group and representation theory we arrive at the first formal definition of disentangled representations. Our new definition is in agreement with many of the current intuitions about disentangling, while also providing principled resolutions to a number of previous points of contention. While this work focuses on formally defining disentangling -- as opposed to solving the learning problem -- we believe that the shift in perspective to studying data transformations can stimulate the development of better representation learning algorithms. 
\end{abstract}

\section{Introduction}
\label{sec_intro}
Recent years have seen significant progress in machine learning (ML), in particular in terms of supervised \cite{Hu_etal_2018, Krizhevsky_etal_2012, Szegedy_etal_2015, He_etal_2016} and reinforcement learning \cite{Mnih_etal_2015, Mnih_etal_2016, Jaderberg_etal_2017, Hessel_etal_2017, Espeholt_etal_2018, Silver_etal_2016}. Many of the best performing algorithms, however, often suffer from poor data efficiency. Furthermore, their performance often lacks the same level of robustness and generalisability that is characteristic of biological intelligence \cite{Lake_etal_2016, Garnelo_etal_2016, Marcus_2018}. 

A long standing idea in ML is that such shortcomings can be reduced by introducing certain inductive biases into the model architecture that reflect the structure of the underlying data \cite{Botvinick_etal_2015, Gens_Domingos_2014, Cohen_Welling_2016, Hinton_etal_2011}. One of the most impactful demonstration of this idea, which caused a step change in machine vision, is the convolutional neural network, in which the translation symmetry characteristic of visual observations is hard wired into the network architecture through the convolution operator \cite{LeCun_etal_1989}. 

An alternative to hard wiring inductive biases into the network architecture is to instead learn a representation that is faithful to the underlying data structure \cite{tishby2000information, Achille_and_Soatto_2018, achille2018information, alemi2016deep, Alemi_etal_2018, Anselmi_etal_2016, Soatto_2010}. This is the motivation for the work on disentangled representation learning, which differs from other dimensionality reduction approaches through its explicit aim to learn a representation that axis aligns with the generative factors of the data \cite{Bengio_etal_2013, Schmidhuber_1992, DiCarlo_Cox_2007}. Note, however, that currently there is little agreement in the field on many aspects of disentangled representations \cite{Locatello_etal_2018}, including what constitutes the data generative factors, whether each data generative factor should be represented with a single or multiple dimensions, and whether the representation should have a unique axis alignment. The lack of a formal definition for disentangled representations makes it hard to evaluate new approaches and measure progress in the field.

Our work attempts to bridge this gap. In particular, we use tools from mathematics and physics to show that, under certain assumptions, a formal connection can be made between the symmetry transformations characteristic of our world and disentangled representations. In particular, our argument is based on the observation that many natural transformations will change certain aspects of the world state, while keeping other aspects unchanged (or invariant). Such transformations are called symmetry transformations, and they can be described and characterised using group and representation theories in mathematics. In particular, these approaches can be used to define a set of constraints on the decomposition of a vector space into independent subspaces to ensure that the vector space is reflective of the underlying structure of the corresponding symmetry group. We apply these insights to the vector space of the model representations and, through that, arrive at the first principled definition of a \emph{disentangled representation}. Intuitively, we define a vector representation as disentangled, if it can be decomposed into a number of subspaces, each one of which is compatible with, and can be transformed independently by a unique symmetry transformation.

Note that this paper only aims to make a theoretical contribution and does not provide a recipe for a general algorithmic solution to disentangled representation learning. It builds a framework to establish a formal connection between symmetry groups and vector representations, which in turn helps resolve many outstanding points of contention surrounding disentangled representation learning. For example, our insights can elucidate answers to questions like what are the ``data generative factors'', which factors should in principle be possible to disentangle  (and what form their representations may take), should each generative factor correspond to a single or multiple latent dimensions, and should a disentangled representation of a particular dataset have a unique basis (up to a permutation of axes). We hope that by gaining a better understanding of what disentangling is and what it is not, faster progress can be made in devising more robust and scalable approaches to disentangled representation learning. 

The rest of the paper is organised as follows. We start by reviewing some of the insights from modern physics on importance of symmetry transformations in our world (Sec.~\ref{sec_physics}). We then present a high-level overview of our perspective on how the connections between symmetry transformations and vector representations can be used to define disentangled representations (Sec.~\ref{sec_high_level_overview}), before giving a brief history of research on disentangled representations and related work on symmetries in machine perception (Sec.~\ref{related_work}). We then provide a formal definition of the high-level definitions from Sec.~\ref{sec_high_level_overview} using the mathematical formalism of group and representation theories (Secs.~\ref{sec_defining_nonlinear_disentangled_representations}-\ref{sec_defining_linear_disentangled_representations}), and discuss the consequences of our definition in the context of the current beliefs about disentangled representations in Sec.~\ref{sec_connections_to_disentanglement}. We start with a non-technical overview to ensure that the readers are introduced to the big picture first and can use the built intuitions to navigate the details of the mathematical formalism that follows. Non-mathematically inclined readers can safely skip Secs.~\ref{sec_defining_nonlinear_disentangled_representations}-\ref{sec_defining_linear_disentangled_representations} and instead concentrate on Secs.~\ref{sec_high_level_overview} and \ref{sec_connections_to_disentanglement}.

\section{Our symmetrical world}
\label{sec_physics}

At the basis of our reasoning lies the assumption that the world dynamics can be described in terms of symmetry transformations that change certain aspects of the world state while keeping other aspects unchanged. This assumption is inspired by the profound role of symmetries in physics. Physics aims to build useful predictive models of our world at different levels of abstraction (e.g. from quantum physics, to electromagnetism, to thermodynamics, to classical mechanics). Symmetry transformations are prevalent at every level of abstraction and ``it is only slightly overstating the case to say that physics is the study of symmetry'' \cite{Anderson_1972}.

Intuitively, a symmetry of an object is a \emph{transformation} that leaves certain properties of the object \emph{invariant}. For example, translation and rotation are symmetries of objects -- an apple is still an apple whether it is placed on a table or in a bag, and whether it rolls on its side or remains upright. Note that the word ``object'' is used in the broadest possible sense -- it could be a mathematical object. For example, the subspace of a unit circle defined by the function $x^2+y^2 \leq 1$ in $\mathbb{R}^2$ remains invariant under the symmetry transformation of axis permutation. 

The study of symmetries in physics in its modern form originates with Noether's Theorem \cite{Noether_1915}, which proved that every conservation law is grounded in a corresponding continuous symmetry transformation. For example, the conservation of energy arises from the time translation symmetry, the conservation of momentum arises from the space translation symmetry, and the conservation of angular momentum arises due to the rotational symmetry. Since then, the study of symmetries in physics has allowed for extraordinary generalisation of prediction to new domains. For example, Gell-Mann \cite{Gell_Mann_1962} predicted the existence of a particle called $\Omega^-$ in 1962 based on the Eightfold Way theory of organising subatomic hadrons according to the flavour symmetries of its constituent quarks. This particle was indeed observed two years later \cite{Barnes_1964}.

Ever since the Noether's Theorem, the study of symmetries has been used to \emph{unify} the existing branches of physics (e.g. the unification of the fields of electricity and magnetism into electromagnetism was implicitly done by unifying the symmetry transformations of the two systems); to \emph{categorise} the known physical objects, (e.g. crystals and elementary particles); and to \emph{discover} new physical objects by attempting to find an entity that fills a vacant place in a symmetry-based classification scheme (e.g. the $\Omega^-$ particle). Modern physics underwent a paradigm shift -- there was a realisation that studying the symmetries of a system can help uncover many new properties of the system itself -- and the emphasis in theoretical physics changed from studying \emph{objects} to studying \emph{transformations}. A similar change in focus in artificial intelligence may help move beyond some of the current limitations of deep learning systems.

While Noether's Theorem is a particularly beautiful and mathematically precise instance of symmetry in the natural world, we should note that there are numerous other instances where a close study of structure-preserving transformations have led to the discovery of new concepts and prediction of new phenomena. For instance, Mendeleev famously predicted the existence of gallium and germanium by studying gaps in his periodic table of elements \cite{Mendeleev_1871}. The more precise and exact these symmetries are, the more powerful the generalisation is to new domains.

In machine perception, the most powerful generalisation we can hope for is by understanding what properties of the world remain the same when transformed in certain ways. This should lead to rapid generalisation in new settings, such as recognising a new object as an object because it demonstrates the same symmetries already recognised in other objects. In scene understanding, these transformations include translations, rotations and changes in object colour, and we will consider these as canonical examples through the rest of the paper. We hope that by providing a formal definition of disentangling we can inspire new algorithms that will allow intelligent systems to display the same rapid generalisation that humans do.

\section{A roadmap to defining disentangled representations}
\label{sec_high_level_overview}
This section introduces our definition of disentangled representations in a non-technical way. One may think of this section as a roadmap to help guide the readers through the technical sections that follow, ensuring that they can see the forest for the trees. Note that we omit many important details here (e.g. what needs to hold in order for a set of transformations to be considered a group), so the reader is strongly encouraged to read the technical sections with care.

As discussed previously, symmetry transformations are ubiquitous in our world (Sec.~\ref{sec_physics}), and many approaches have tried to bring ideas from group theory to machine learning in the past (see Sec.~\ref{related_work} for an overview). However, none of these approaches so far have formally connected symmetry groups to disentangled representations. How can this connection be made?

\begin{figure}[h!]
 \centering
 \includegraphics[width = 0.7\textwidth]{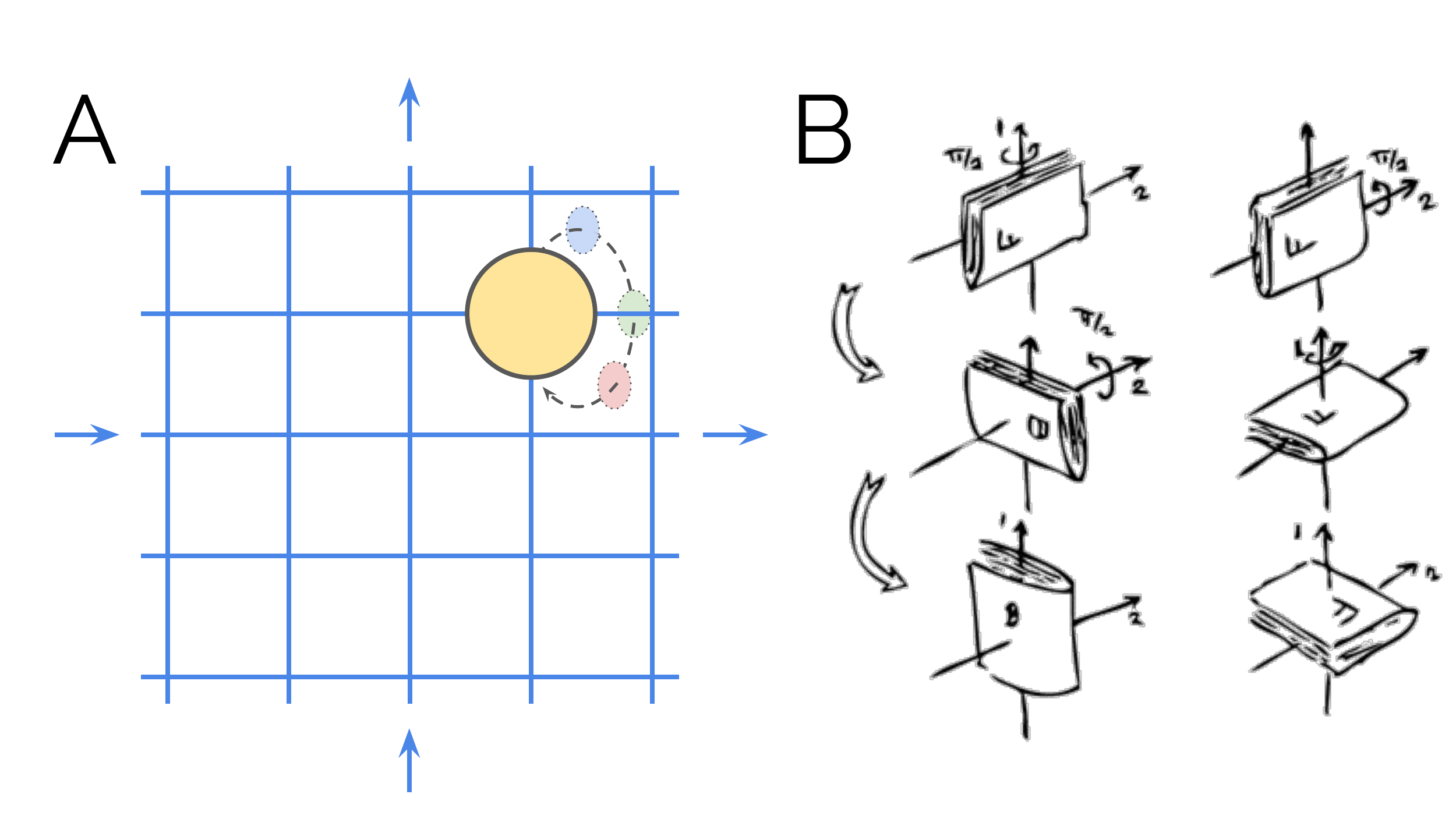}
 \vspace{-6pt}
 \caption{\textbf{A}: an example grid world where the object can move horizontally or vertically, as well as change colour. Moving beyond the edge of the grid transports the object to the opposite side of the grid.  \textbf{B}: an example of non-commutativity of 3D rotations. Rotating by $90^{\circ}$ along axis 1 then axis 2 is different than rotating by the same amount and around the same axes but in the opposite order.}
 \label{fig_example_worlds}
\end{figure}

Let us consider a concrete example. Imagine a grid world with a single object, which can move around in four directions (up, down, left, right), and change colour by taking discrete steps on a circular hue axis. Whenever the object steps beyond the boundary of the grid world, it is placed at the opposite end (e.g. stepping up at the top of the grid places the object at the bottom of the grid) (see Fig.~\ref{fig_example_worlds}A).

Horizontal and vertical translations and colour changes are the symmetry transformations of such a world, since none of these transformations affect the identity of the object. The set of these transformations makes up a symmetry group, and the effects of these transformations on the state of the world are called the \emph{actions} of the symmetry group on the world state. 

Note that the horizontal translation action changes the horizontal position of the object, but not its vertical position or colour, the vertical translation action changes the vertical position, but not the horizontal position or colour, and finally the colour change action changes the object's colour, but not its position. We define such actions, which change a certain aspect of the world state, while keeping others fixed, as \emph{disentangled group actions}. Furthermore, we can say that the symmetry group of this world can be decomposed into three separate symmetry \emph{subgroups}: one that affects the horizontal translations, another affecting vertical translations and the last one affecting colour changes.

So far we have only considered the transformations of the abstract world \emph{states}. However, in order to define disentangled representations, we need to bring the \emph{observations} of such abstract states into the picture. We do so by assuming that there exists a certain generative process that produces a dataset of observations from a set of world states. This process may have many different instantiations depending on the perceptual system of the observer. Without the loss of generality, in this paper we assume that the generative process produces pixel observations. 

The pixel observations then form an input to a model that produces a representation (the inference process). We consider a composition of the generative and the inference processes, going from the abstract world states to the model representation, and argue that in some situations it may be possible to find such a composite mapping, so that it also preserves the abstract disentangled group action. In other words, we want to find a mapping between the disentangled group action in the abstract state space and the transformations in the vector space of representations, so that the latter reflect the structure of the former. If such a mapping exists, then it produces a \emph{disentangled representation}. Hence:

\vspace{5mm}
    \emph{A vector representation is called a \textbf{disentangled representation} with respect to a particular decomposition of a symmetry group into subgroups, if it decomposes into independent subspaces, where each subspace is affected by the action of a single subgroup, and the actions of all other subgroups leave the subspace unaffected.}
\vspace{5mm}

This definition of disentangled representations is very general -- it does not assume any particular dimensionality or basis for each subspace. Furthermore, there are no constraints on the nature of the disentangled group action on each subspace, as long as these actions affect the model representation in a way that is equivalent to how they affect the abstract world states. However, it is important to note that we define disentangled representations \emph{with respect to a particular decomposition of a symmetry group into subgroups}. This aspect of our definition will be discussed in more detail later in this section. 

A representation disentangled by the definition above will have the \emph{compositional} property often found to be desirable. For example, subsequent tasks may safely ignore certain parts of such a representation to form abstractions \cite{Higgins_etal_2018, Laversanne-Finot_etal_2018}, or use novel recombinations of values within the subspaces for improved generalisation \cite{Higgins_etal_2017b}. However, a disentangled representation may be even more useful if the additional \emph{linearity} property is imposed on group actions in its vector space. For example, if the colour and position changes can be implemented using linear transformation matrices applied to the vector space of a disentangled representation, then an agent equipped with such a representation could learn a transition model of the grid world much more efficiently than if the transitions (i.e. group actions) were non-linear. Hence, we also propose a definition for a \emph{linear disentangled representation}:

\vspace{5mm}
    \emph{A vector representation is called a \textbf{linear disentangled representation} with respect to a particular decomposition of a symmetry group into subgroups, if it is a disentangled representation with respect to the same group decomposition, and the actions of all the subgroups on their corresponding subspaces are linear.}
\vspace{5mm}

What would be a disentangled representation of our example grid world? The vector space of such a representation would be a concatenation of three independent multi-dimensional vector subspaces, such that, for example, a change in colour only affects the ``colour subspace'', but not the ``position x'' and ``position y'' subspaces. The changes along each of the subspaces in the representation may be implemented by an arbitrary mapping. If this mapping is linear, then the representation is a \emph{linear disentangled representation}  (see Secs.~\ref{sec_worked_nonlinear_example} and \ref{sec_worked_linear_example} for more details). Finally, each subspace may have an arbitrary suitable basis. For example, the ``colour subspace'' may have a multi-dimensional RGB or a single-dimensional hue basis. Furthermore, we do not require the ``position x'' and ``position y'' subspaces to actually align with the Cartesian coordinates. Any rotation of the axes is also acceptable. An \emph{entangled representation} of the same grid world will have subspaces that are affected by the actions of more than one subgroup. 

Note that we define disentangled representations with respect to a particular decomposition of a symmetry group into subgroups. This has two important consequences. First, it helps shed light on some current areas of disagreement, where it is not clear whether a representation for a particular dataset could be disentangled in principle. For example, consider rotations of an object in 3D. These form a symmetry group, since they do not affect the object identity. Intuitively, we might think that a representation of such a group could be disentangled into rotations about the x, y and z axes. However, changes in rotation around the different axes will affect each other, since rotations in 3D are not commutable. For example, rotating the object by $90^\circ$ first around the x and then the y axis will not be the same as rotating by the same angle and around the same axes but in the opposite order (see Fig.~\ref{fig_example_worlds}B). In other words, when we consider the symmetry group of 3D rotations, we find that rotations about the different axes do not commute with one another, and hence the group of symmetries does not decompose in the way that we might intuitively hope. This means that the resulting representation should contain a single subspace affected by rotations around all three axes, which a priori dooms the prospect of disentangling these factors.

Another consequence of defining disentangled representations with respect to a particular decomposition of a symmetry group is that symmetry groups may have multiple possible subgroup decompositions, and not all group decompositions may lead to disentangled representations useful for subsequent tasks. For example, one of the decompositions of the symmetry group acting on our grid world example may be the one we have already discussed: three subgroups representing changes in colour, position x and position y. However, for some instantiations of the group, it may also decompose into more than three subgroups, where each of the colour and each of the two position changes are themselves decomposed into multiple subgroups. This is not particularly problematic, since subsequent tasks may pick out the subset of the most useful disentangled subspaces. However, another possibility is that the group may decompose into subgroups that mix colour and position changes, which is clearly not desirable. This may happen because by themselves abstract groups often do not provide a way to identify useful group decompositions. 

Saying this, we believe that useful group decompositions can be discovered empirically through \emph{active perception}. By acting in the world, agents should be able to discover which aspects of the world remain invariant under various transformations. For example, gravity based manipulations can be used to decompose a group of 3D translations into subgroups of translations in the vertical and horizontal planes. Similar arguments were also made in \cite{Soatto_2010}. This also relates to the recent work by \cite{Locatello_etal_2018}, who proved that unsupervised learning of disentangled representations is impossible unless certain biases are introduced into the model architecture or the data. The latter can be done through active perception or causal manipulations of the world \cite{Suter_etal_2018}. 

To summarise, for a given symmetry group, it may be the case that many different subgroup decompositions are possible. We assume that the structure of the world dictates a certain \emph{natural} decomposition. In our example, we would regard a decomposition into position and colour as natural. A decomposition in which position and colour remain mixed is possible, but not natural. One problem that a learning algorithm needs to solve is to find natural decompositions that reflect the structure of the world, rather than the unnatural ones. In earlier approaches, it is usually assumed that natural decompositions can be found on the basis of statistical independence structure found in the data, which is assumed to be i.i.d. We hope that our approach lends itself better to an active learning approach, in which the data cannot be assumed to be i.i.d.

Note that the question of finding a useful group decomposition is a separate question from defining disentangled representations, and we leave it for future work. Hence, in the rest of the paper we will assume that a useful group decomposition is given. Regardless of which particular group decomposition a disentangled representation is defined in terms of, the insights on the structure of the resulting representation will still hold; this provides a principled resolution to many current points of disagreement about disentangled representations. We discuss the consequences of our definition in the context of the current beliefs about disentangled representations in more detail in Sec.~\ref{sec_connections_to_disentanglement}.

In the rest of the paper we will illustrate our ideas using toy idealised examples. Our insights, however, are generalisable to the more realistic environments, but we leave the question of how to empirically learn disentangled representations to future work.

\section{Related work}
\label{related_work}
There is a long history of research in psychology and artificial intelligence that attempts to formalise the idea that different factors of variation in the world can be recovered from raw perceptual inputs. It was recognised early on in the study of vision that certain transformations would leave object identity invariant, that there are many such transformations which are independent of one another, and that understanding how the visual system forms representations invariant to these transformations is critical to understanding perception. For instance, J. J. Gibson in his classic work on the ecological approach to visual perception \cite{Gibson_1979} wrote:

\begin{quote}
Four kinds of invariants have been postulated: those that underlie change of illumination, those that underlie change of the point of observation, those that underlie overlapping samples, and those that underlie a local disturbance of structure.
\end{quote}
Following Gibson's work, psychologists almost immediately picked up on the mathematical framework of group theory as a way of formalising the notion of invariants and symmetry \cite{Dodwell_1983}, but progress in applying this framework usefully in computer vision would take many decades.

Much of the research on perception and representation learning, especially in object recognition, has followed Gibson and his contemporaries in emphasising features which are {\em invariant} to transformations like pose or illumination \cite{Lowe_1999, Dalal_and_Triggs_2005, Sundaramoorthi_etal_2009, Krizhevsky_etal_2012}. In this framework, transformations are considered nuisance variables to be thrown away. However, other researchers have advocated an approach to representation learning which preserves information about these transformations - in other words, representations that are {\em equivariant} to transformations instead of invariant \cite{Hinton_etal_2012}. In the equivariant approach to perception, certain subsets of features may be invariant to specific transformations, but the overall representation will still preserve all information. Especially in an unsupervised setting, where the representations may be used for many different tasks, and it is not known ahead of time which transformations are ``nuisances'', an equivariant approach to perception seems more appropriate. Work on learning disentangled representations falls into this line of research.

Significant work in machine perception, both invariant and equivariant, has tried to use the machinery of group theory to quantify natural symmetries. While classic work showed that there are no general-purpose features for 3D point clouds that are invariant to changes in viewpoint \cite{Burns_etal_1992}, later work on more realistic models of images showed that features simultaneously invariant to changes in illumination and viewpoint do in fact exist \cite{Sundaramoorthi_etal_2009}. These two transformations are invertible, and hence can be described as elements of a group. Other transformations like occlusion can be made invertible if the viewer is able to interact with the scene. This is the basis of the theory of {\em actionable information}, which holds that a complete theory of visual perception needs to treat image acquisition as an active process of information gathering \cite{Soatto_2010}. The term ``disentangling" is used frequently in this line of research, including later work in a deep learning framework \cite{Achille_and_Soatto_2018}, but in this context it is meant more in the sense of disentangling (and throwing away) nuisance variables and preserving task-relevant information. This is not the sense in which we use the term disentangling.

Group theory also forms the foundation of Anselmi et al's theory of visual recognition with hierarchical models \cite{Anselmi_etal_2016}. In their framework, an ``object'' is defined as the orbit of a prototypical object under all symmetries from a particular group. The goal of hierarchical visual models is then to find an {\em invariant signature} which maps all elements of this group orbit to a single descriptor, while maintaining a large difference between descriptors for different objects. While group transformations are used here to define the space of nuisance parameters, there is little discussion of the structure of the groups themselves. Our notion of disentangling in this paper is primarily concerned with the way that the symmetry group of the invariant transformations decomposes into simpler subgroups for each separate transformation, a topic not addressed in \cite{Anselmi_etal_2016}.

The idea of {\em learning} symmetry transformations from data has appeared many times, mostly in the context of continuous symmetries like translation and rotation. Earlier work in this area mostly focused on learning transformations in observation space, such as translation and in-plane rotation of objects \cite{Rao_Ruderman_1999, Sohl-Dickstein_etal_2010, Cohen_Welling_2014}. Later work extended this to general 3D rotations \cite{Cohen_Welling_2015}, but did not learn to disentangle this rotation from other factors of variation. Though it does not touch on all aspects of the theory outlined here, we believe the work in \cite{Cohen_Welling_2014, Cohen_Welling_2015} is the most closely related work to our theory of disentangling for its emphasis on irreducible group representations.

Our definition of disentangling relies heavily on group theory, and it is worth mentioning many other ways ideas from group theory have influenced machine learning, such as orthonormality constraints on network weights \cite{Fiori_2002, Cisse_etal_2017}, extensions to convolutional network architectures that build in full affine invariance rather than just translation invariance \cite{Gens_Domingos_2014}, and discovering abstractions by partitioning sets \cite{Yu_etal_2018}. However, none of these explicitly address the way disentangled representations can be learned or defined.

Separate from much of this work on group theory, a significant part of the deep learning community in recent years has become focused on disentangling as a high-level goal of unsupervised learning, and tried to build models that can automatically discover the disentangled factors of variation in data. The idea of learning a distributed code where each feature would correspond to a relevant factor of variation dates back at least to Schmidhuber \cite{Schmidhuber_1992}, though the use of the term ``disentangling" for this property of a model appeared much later \cite{Bengio_2009}.

Most initial attempts to learn disentangled representations required a priori knowledge of the data generative factors \cite{Hinton_etal_2011, Rippel_Adams_2013, Reed_etal_2014, Zhu14, Yang_etal_2015, Goroshin_etal_2015, Kulkarni_etal_2015, Cheung15, Whitney_etal_2016, Karaletsos_etal_2016}. This, however, is unrealistic in most real world scenarios. A number of purely unsupervised approaches to disentangled factor learning have been proposed \cite{Schmidhuber_1992, Desjardins_etal_2012, Tang13}, however they did not tend to scale well beyond toy datasets. Recently, this shortcoming was overcome by two new approaches to unsupervised disentangled representation learning developed concurrently and independently: InfoGAN \cite{Chen_etal_2016}, based on Generative Adversarial Networks (GANs) \cite{Goodfellow_etal_2014}, and \betavae \cite{Higgins_etal_2017}, based on Variational Autoencoders (VAEs) \cite{Kingma_Welling_2014, Rezende_etal_2014}. InfoGAN was found to suffer from training instabilities and to perform worse in terms of disentangling \cite{Higgins_etal_2017, Kim_Mnih_2018}, often failing to discover many of the data generative factors. Hence, the majority of the most recent techniques have built upon the \betavae approach \cite{Burgess_etal_2017, Kim_Mnih_2018, Chen_etal_2018, Kumar_etal_2017, Ansari_Soh_2018, Esmaeili_etal_2018}, improving on the poor disentanglement/reconstruction trade-off of the original.

To date, most work on disentangling has relied on comparison with human intuition for evaluation, though a number of metrics have been proposed in the case where the ground truth factors are known \cite{Eastwood_Williams_2018, Ridgeway_Mozer_2018, Kumar_etal_2017, Kim_Mnih_2018, Locatello_etal_2018, Suter_etal_2018}. The lack of clear definition is the main motivation for this work, and the relationship between our definition and prior evaluation metrics is discussed in Sec.~\ref{sec_connections_to_disentanglement}.

\section{A formal definition of disentangled representations}
\label{sec_defining_nonlinear_disentangled_representations}

This section provides a formal definition of \emph{disentangled representations}. It assumes a certain degree of familiarity with group theory. Readers unfamiliar with the preliminaries outlined below are referred to Appendix~\ref{sec_group_theory} for a review of the elementary concepts of group theory.

\vspace{5mm}
\textbf{Preliminaries (Appendix~\ref{sec_group_theory}):} group $(G)$, binary operator $(\circ: G \times G \rightarrow G)$, group decomposition into a direct product of subgroups $(G = G_1 \times G_2)$, set $X$, group action on $X$ $(\cdot : G \times X \rightarrow X)$.

\vspace{5mm}

We start by defining a \emph{disentangled group action}, before using it to define a \emph{disentangled representation}.

\subsection{Disentangled group action}
\label{sec_disentangled_action}
Suppose that we have a group action $\cdot: G \times X \rightarrow X$, and that the group $G$ decomposes as a direct product $G = G_1 \times G_2$. We are going to refer to the action of the full group as $\cdot$, and the actions of each subgroup as $\cdot_i$.  Then we propose the following definition: the action is \emph{disentangled} (with respect to the decomposition of $G$) if there is a decomposition $X = X_1 \times X_2$, and actions $\cdot_i: G_i \times X_i \rightarrow X_i, i \in \{1,2\}$ such that 
\begin{equation} \label{eq_disentangled_action}
(g_1, g_2) \cdot (v_1, v_2) = (g_1 \cdot_1 v_1, g_2 \cdot_2 v_2)
\end{equation}
In particular this says that an element of $G_1$ acts on $X_1$ but leaves $X_2$ fixed, and vice versa.

If $X$ is a space with additional structure, such as a vector space or a topological space, then we may be interested in actions that preserve that structure (linear or continuous actions). We remark that in that case the subspace actions also preserve this structure. For example, if $X$ is a vector space, and the action is linear, then the subspace actions are linear too.

When $X$ is a vector space -- even if the action of $G$ on $X$ is not linear -- then we write the decomposition instead as $X = X_1 \oplus X_2$. We emphasise that if a basis is given for $X$, we do not require that the decomposition factors $X_1$ and $X_2$ are aligned to this basis.

Our definition extends to group decompositions $G = G_1 \times ... \times G_n$. In that case, we say that the action is disentangled (with respect to the decomposition of $G$) if there is a decomposition $X = X_1 \times ... \times X_n$ or $X = X_1 \oplus ... \oplus X_n$ such that each $X_i$ is fixed by (is invariant to) the action of all the $G_j, j \neq i$ and affected only by $G_i$.

\subsection{Disentangled representation}
\label{sec_defining_disentangled_representations}
In this section, we would like to show how the group structure of symmetries of the world can have implications for an agent's internal representations.

Let $W$ be the set of world-states. We suppose that there is a generative process $b: W \rightarrow O$ leading from world-states to observations (these could be pixel, retinal, or any other potentially multi-sensory observations), and an inference process $h: O \rightarrow Z$ leading from observations to an agent's representations. At times we will assume that $Z$ is a vector space (whereas $W$ is assumed only to be a set). We consider the composition $f: W \rightarrow Z$, $f = h \circ b$. 

Suppose also that there is a group $G$ of symmetries acting on $W$ via an action $\cdot: G \times W \rightarrow W$. What we would like is to find a corresponding action $\cdot: G \times Z \rightarrow Z$ so that the symmetry structure of $W$ is reflected in $Z$\footnote{Note that in the rest of the section $\cdot$ may indicate two different things:  actions of $G$ on the abstract states $W$ or  actions of $G$ on the vector space of representations $Z$. The relevant group action should be referred from the context.}.  In other words, we want the action on $Z$ to correspond to the action on $W$. This can be achieved if the following condition is satisfied: 
\begin{equation} \label{eq_equivariant_map}
g \cdot f(w) = f(g \cdot w) \ \ \forall g \in G, w \in W
\end{equation}
This states that the action should commute with $f$. This is the definition of a $G$-morphism or an equivariant map. Hence, $f$ can be called a $G$-morphism or an equivariant map. 

\[ \begin{tikzcd}
G \times W \arrow{r}{\cdot_W} \arrow[swap]{d}{id_G \times f} & W \arrow{d}{f} \\%
G \times Z \arrow[dashed]{r}{\cdot_Z}& Z
\end{tikzcd}
\]

For a given $f: W \rightarrow Z$, there is no guarantee that we can find a compatible action $\cdot: G \times Z \rightarrow Z$ satisfying Eq.~\ref{eq_equivariant_map}. If $f$ is bijective then in fact Eq.~\ref{eq_equivariant_map} can be taken as the definition of the action (setting $z = f(w)$, we have $g \cdot z = f(g \cdot f^{-1}(z))$). If $f$ is injective but not surjective, then this recipe does not tell us how to define the action on the part of $Z$ which is not in the image $f(W)$. However, this does not matter: we do not care about all of $Z$, but only about the action of $G$ on the parts of $Z$ that are mapped to by $f(W)$. If $f$ is not injective, then there exist $w_1$, $w_2$ such that $f(w_1) = f(w_2)$. For example, this may happen because different world states give rise to the same observations, because of occlusion. In theory this can be a problem, however, such non-invertible mappings can be made invertible in practice through active sensing, as discussed in \cite{Soatto_2010}.  

In Sec.~\ref{sec_worked_nonlinear_example}, we give an example of an $f: W \rightarrow Z$ satisfying the equivariance condition. For the remainder of this section, we assume that we have an action $\cdot: G \times Z \rightarrow Z$ and an equivariant map $f: W \rightarrow Z$.

Suppose also that the symmetries of the world decompose as a direct product $G = G_1 \times ... \times G_n$. In Sec.~\ref{sec_disentangled_action}, we defined what it means to say that a group action is disentangled with respect to such a group decomposition. We now propose the following definition: We say that an agent's representation $f: W \rightarrow Z$ is disentangled if the conditions above are satisfied, and the action $\cdot$ on $Z$ is disentangled according to the earlier definition.

In other words, an agent's representation $Z$ is disentangled with respect to the decomposition $G = G_1 \times ... \times G_n$ if
\begin{enumerate}
    \item There is an action $\cdot: G \times Z \rightarrow Z$,
    \item The map $f: W \rightarrow Z$ is equivariant between the actions on $W$ and $Z$, and
    \item There is a decomposition $Z = Z_1 \times ... \times Z_n$ or $Z = Z_1 \oplus ... \oplus Z_n$ such that each $Z_i$ is fixed by the action of all $G_j, j \neq i$ and affected only by $G_i$.
\end{enumerate}

When $Z$ has additional structure (such as a vector space or topological space) then we may consider actions that preserve that structure (linear actions or continuous actions). For vector spaces in particular there is a well-developed theory of \emph{group representations}, which we discuss in more detail in Sec.~\ref{sec_defining_disentangled_representations_linear}.

Note that in the preceding we did not assume that the action of $G$ on $W$ is disentangled. Nevertheless, this is a very natural assumption to make, in which case there would be a decomposition $W = W_1 \times ... \times W_n$ such that the action $G \times W \rightarrow W$ decomposes into sub-actions $G_i \times W_i \rightarrow W_i$, $i \in \{1..n\}$. Although this assumption is not necessary for our argument, it is a natural way to express our intuition that the world has compositional structure that we would like our agent's representation to reflect.

In summary:
\begin{itemize}
    \item Given a world $W$ with symmetries $G$, we have shown how it is possible for an agent's representation $f: W \rightarrow Z$ to induce corresponding symmetries in the agent's representation space $Z$.
    \item We have given a definition of what it means for an agent's representation to be disentangled with respect to a factorisation of the symmetry structure of $W$ into independent factors of variation as $G = G_1 \times ... \times G_n$.
\end{itemize}

\subsection{A worked example of a disentangled representation}
\label{sec_worked_nonlinear_example}

Let us consider what a disentangled representation of our grid world example from Sec.~\ref{sec_high_level_overview} might look like. The gridworld $W$ can be described by a symmetry group $G = G_x \times G_y \times G_c$, where $G_x$ is the set of all translation transformations along the x axis, $G_y$ is the set of all translation transformations along the y axis, and $G_c$ is the set of all colour transformations. Note that $G_x$, $G_y$, $G_c$ are each isomorphic to the cyclic group $C_N$.

\begin{figure}[h!]
 \centering
 \includegraphics[width = 1.0\textwidth]{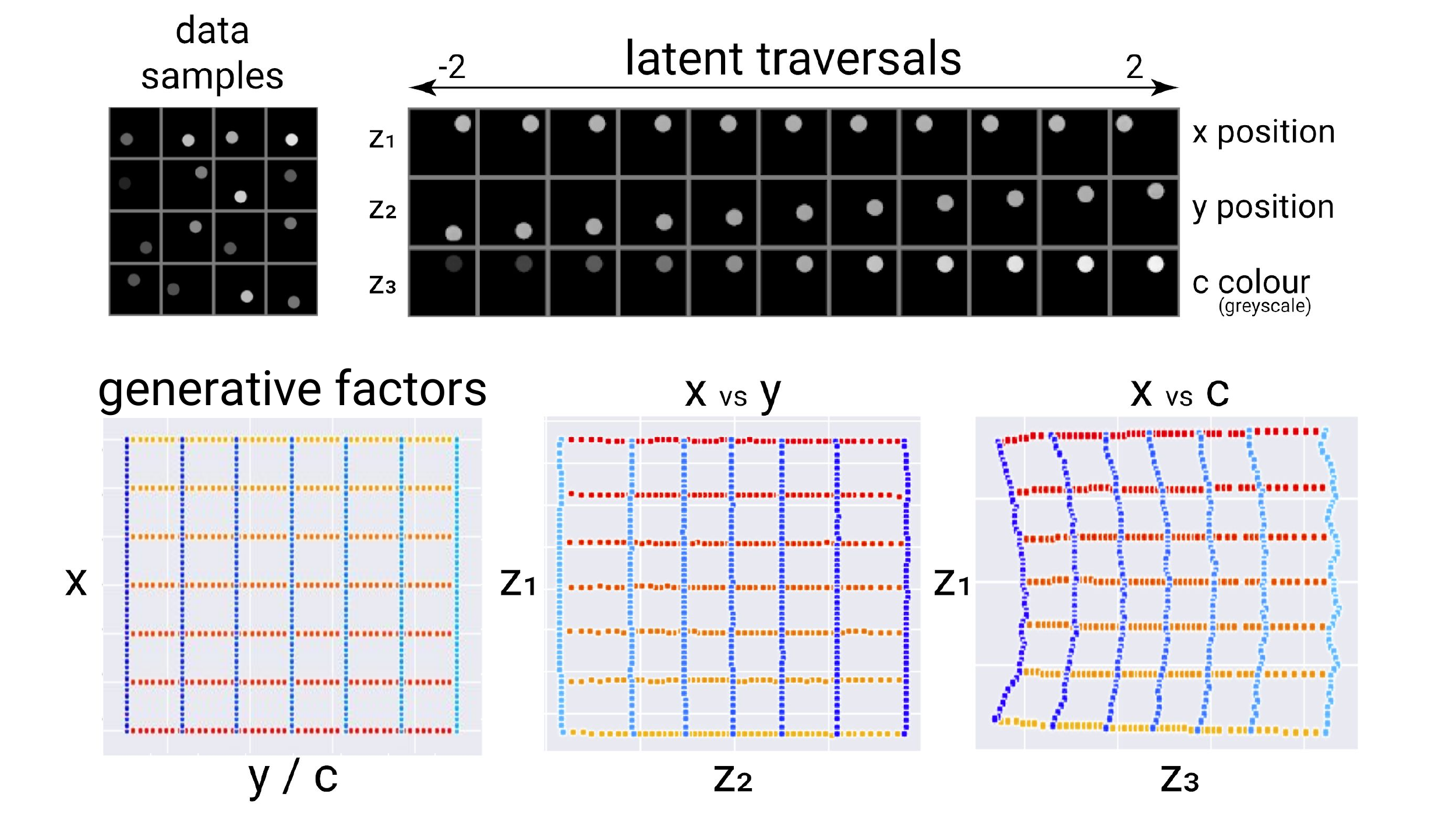}
 \vspace{-6pt}
 \caption{\textbf{Top left}: pixel observations $o \in O$ of world states $w \in W$ under the action of group $G = G_x \times G_y \times G_c$, where $G_x$ stands for a cyclic group of translations along the x coordinate, $G_y$ stands for a cyclic group of translations along the y coordinate, and $G_c$ stands for a cyclic group of translations along the colour axis (in greyscale). \textbf{Top right}: latent traversals for a CCI-VAE \cite{Burgess_etal_2017} model trained on the data distribution. The value of each latent is traversed between $[ -2, 2 ]$ while the other latents are fixed to their inferred value. It can be seen that the model learnt a representation that approximates the action of the group $G$. \textbf{Bottom}: visualisation of the projection $f: O \rightarrow Z$ of certain world states onto the subspaces of the agent's representation spanned by $z_1 \times z_2$ and  $z_1 \times z_3$. Qualitatively $f$ appears bijective, and the operation of group $G$ appears to be approximately preserved (apart from the cyclic aspect) in its representation on $Z$, i.e. $f(\textrm{``move right''} \circ \textrm{``move up''}) \approx f(\textrm{``move right''}) \circledcirc f(\textrm{``move up''})$ suggesting that CCI-VAE learnt a good approximation that matches the action of the group $G$ on $Z$. }
 \label{fig_cc_vae_grids}
\end{figure}

We trained one of the current state of the art disentangling models CCI-VAE \cite{Burgess_etal_2017} on a dataset of observations from such a grid world. Fig.~\ref{fig_cc_vae_grids} shows that the latent representation learnt by the model is an acceptable approximation for the underlying group action of $G = G_x \times G_y \times G_c$. In particular, in this example the generative process $b: W \rightarrow O$ is the graphics engine we used to generate the dataset, and the inference process $h: O \rightarrow Z$ is the encoder of CCI-VAE. We can see that $f$ is indeed an approximately equivariant map as required by our definition $f(x, y, c) \approx (\lambda_x x, \lambda_y y, \lambda_c c)$, where $\lambda_i \in \mathbb{R}, \ \ i \in \{x, y, c \}$ and so we may take $Z = Z_1 \times Z_2 \times Z_3$ as the coordinate axes. Each subgroup acts independently on its corresponding subspace in the latent representation, and a lot of the group structure is preserved (e.g. the commutativity of the group actions). Note, however, that the group action on the learnt representation space is a translation and hence is not linear. Furthermore, the cyclical nature of the group has been lost. An alternative representation with a linear group action will be discussed in Sec.~\ref{sec_worked_linear_example}.

Note that, as discussed in Sec.~\ref{sec_high_level_overview}, the symmetry group $G$ may have many possible decompositions. The representation discussed above is disentangled with respect to a particular group decomposition $G = G_x \times G_y \times G_c$. However, cyclic groups $C_N$ may decompose into smaller subgroups for all non-prime $N$. For example, for $N=4$, the following group decompositon may also be possible $G= (G_x^1 \times G_x^2) \times (G_y^1 \times G_y^2) \times (G_c^1 \times G_c^2)$. A representation disentangled with respect to this new decomposition would then have six invariant subspaces, and three of them would be redundant. Another possibility is to decompose the group into $G = G_p \times G_c$, where $G_p$ is the subgroup of all positional changes. A representation disentangled with respect to this decomposition would have two invariant subspaces, and the basis for the ``position'' subspace may be any rotation of the Cartesian axes. Note that the question of learning a useful group decomposition is beyond the scope of this paper, however we believe that active learning may play an important part in this regard. Alternatively a search heuristic like the one proposed in \cite{Yu_etal_2018} may be useful.

Now let us also consider the example of 3D rotations from Sec.~\ref{sec_high_level_overview}. Rotations in 3D are isomorphic to the special orthogonal group $SO(3)$. Intuitively we might think that an agent's representation could disentangle rotations about the x, y and z axes. $SO(3)$ indeed has three subgroups $G_x$, $G_y$ and $G_z$ consisting of rotations about the x, y and z axes. Moreover, the intersection of any two of these subgroups is $\{e\}$, and taken together, the subgroups generate the group. However, the problem is that elements of these subgroups do not commute with one another, and so $SO(3)$ cannot be written as a direct product of these subgroups. While the subgroups express factors of variation, they are not independent. Our definition of disentangled representation rules out disentangling along these lines, which our representation indeed fails to capture. To be clear, this does not rule out that object rotations can be disentangled from other symmetries, such as colour changes. In that case we would be interested in the group $G = SO(3) \times G_c$, where $G_c$ is the subgroup of colour change symmetries. This is a direct product, since presumably rotations are independent from colour changes.

\section{A formal definition of linear disentangled representations}
\label{sec_defining_linear_disentangled_representations}

This section builds on the material from Sec.~\ref{sec_defining_nonlinear_disentangled_representations} to provide a formal definition of \emph{linear disentangled representations}. Our definition of disentangled representations from the previous section does not make any assumptions on what form the group action should take when acting on the relevant disentangled vector subspace. However, many subsequent tasks may benefit from a disentangled representation where the group actions transform their corresponding disentangled subspace linearly. This section follows a similar pattern to Sec.~\ref{sec_defining_nonlinear_disentangled_representations} in connecting abstract group actions to the group actions on the vector space of representations. However, we now add an additional linearity constraint to our construction. This requires a certain degree of familiarity with group representation theory. Readers unfamiliar with the preliminaries outlined below are referred to Appendix~\ref{sec_representation_theory} for a review the elementary concepts of group representation theory. 

\vspace{5mm}
\textbf{Preliminaries (Appendix~\ref{sec_representation_theory}):} vector space $(V)$, tensor product of vector spaces $(V \otimes W)$, group action on a vector space $(\cdot : G \times V \rightarrow V)$, group representation $(\rho: G \rightarrow GL(V))$, direct sum of representations $(\rho_1 \oplus \rho_2)$, irreducible representation, tensor product representation $(\rho_1 \otimes \rho_2: G \rightarrow GL(V \otimes W))$.

If a group is a direct product $G = G_1 \times G_2$, then the irreducible representations of $G$ are precisely the representations $\rho_1 \otimes \rho_2$, for $\rho_1, \rho_2$ irreducible representations of $G_1, G_2$ respectively (see Lemma~22.6 in \cite{Teleman_2005}).
\vspace{5mm}

We start by defining a \emph{disentangled group representations}, before using it to define a \emph{linear disentangled representation}.

\subsection{Disentangled group representation}
\label{sec_disentangled_group_representation}
We say that a group representation $\rho: G \rightarrow GL(V)$ is \emph{linearly disentangled} with respect to the group decomposition $G = G_1 \times G_2$ if there exists a decomposition $V = V_1 \oplus V_2$ and representations $\rho_i: G_i \rightarrow GL(V_i), \, i \in \{1,2\}$ such that $\rho = \rho_1 \oplus \rho_2$, that is
$$\rho(g_1, g_2)(v_1, v_2) = (\rho_1(g_1)v_1, \rho_2(g_2)v_2)$$
(This is just Eq.~\ref{eq_disentangled_action} rewritten in terms of group representations. Compared to that equation, the additional assumption here is that the action $\rho$ is linear.)

In terms of irreducible representations, we know that $V$ can be decomposed as $V = V_1 \oplus ... \oplus V_m$ where each factor $V_j$ is a minimal invariant subspace under the action of $G$. We also know that for each of these irreducible subspaces $V_j$, the subrepresentation $\rho$ restricted to the subspace has the form $\rho_1 \otimes \rho_2$, for $\rho_i$ an irreducible representation of $G_i$. What the definition of a disentangled representation says is that in each $V_j$, at most one of the $\rho_i$ is a non-trivial representation, so that $V_j$ is a representation either of $G_1$ or of $G_2$, not both. In that case, we may gather together the $V_j$ that are representations of $G_1$ as the first factor, and the $V_j$ that are representations of $G_2$ as the second factor, to give a disentangled representation.

All of the above extends in an obvious way to decompositions $G = G_1 \times ... \times G_n$. In that case, the irreducible representations of $G$ will be of the form $\rho_1 \otimes ... \otimes \rho_n$, where each $\rho_i, i \in \{1..n\}$ is an irreducible representation of $G_i$, and any representation $\rho$ of $G$ will be a direct sum of such representations. Then the representation $\rho$ is disentangled if in each such factor $\rho_1 \otimes ... \otimes \rho_n$, at most one of the $\rho_i$ is a non-trivial representation.
\subsection{Linear disentangled representation}
\label{sec_defining_disentangled_representations_linear}

Previously we defined what it means to say that an agent's representation is disentangled, using the concept of a disentangled group action. If additionally, the group action is linear, and hence a group representation, then we may define the concept of a linearly disentangled (agent) representation analogously. Linearity is a strong constraint, with the result that we can say much more about group representations than we could about group actions.

As before (see Sec.~\ref{sec_defining_disentangled_representations}), we assume a set $W$ of world-states, a group $G$ acting on $W$ via $\cdot: G \times W \rightarrow W$, a generative process $b: W \rightarrow O$ leading from world-states to observations, an inference process $h: O \rightarrow Z$ leading from observations to an agent's representations, and the composition $f: W \rightarrow Z$, $f = h \circ b$. As before, we are interested in finding an action $\cdot: G \times Z \rightarrow Z$, such that $f$ is equivariant between the two actions $\cdot: G \times W \rightarrow W$ and  $\cdot: G \times Z \rightarrow Z$. However, in this case we impose the additional constraint that the action on $Z$ should be linear, so that we can also view it as a group representation $\rho: G \rightarrow GL(Z)$

Then a \emph{linear disentangled representation} is just an $f: W \rightarrow Z$ that admits such an action, and is disentangled with respect to a decomposition $G = G_1 \times G_2$. From the previous section, we see that this means that there is a decomposition $Z = Z_1 \oplus ... \oplus Z_m$, where each factor $Z_j$ is a representation either of $G_1$ or of $G_2$ alone. Specifically, what is excluded is factors of the form $\rho_1 \otimes \rho_2$, where $\rho_i$ is an action of $G_i$, unless at most one of the $\rho_i$ is a non-trivial representation.

\subsection{A worked example of a linear disentangled representation}
\label{sec_worked_linear_example}

Let us consider what a linear disentangled representation of our grid world example from Sec.~\ref{sec_high_level_overview} might look like. From Sec.~\ref{sec_worked_nonlinear_example} we know that the symmetry group acting on the gridworld $W$ can be decomposed into a direct product $G = G_x \times G_y \times G_c$, where $G_x$ is the set of all translation transformations along the x axis, $G_y$ is the set of all translation transformations along the y axis, and $G_c$ is the set of all colour transformations. The set of abstract world states for such a world will be the Cartesian product of all possible positions and colours:

\[
W = \{ (x_1, y_1, c_1), (x_1, y_1, c_2), ..., (x_N, y_N, c_N) \}
\]
where $|W| = N^3$.

In this simple example, we can easily exhibit an equivariant map $f: W \rightarrow Z$, as follows. Let $Z = \mathbb{C}^3$, which we identify with $\mathbb{R}^6$. Then define:
$$f(x, y, c) = (e^{2\pi ix/N}, e^{2\pi iy/N}, e^{2\pi ic/N})$$
Intuitively, this seems like a good representation to learn. We now show that there is a linear action (a group representation) of $G$ on $Z$, and that it is disentangled.

Suppose that $g_x$, $g_y$, and $g_c$ are generators of the subgroups $G_x$, $G_y$ and $G_c$ respectively. Then we may define $\rho: G \times Z \rightarrow Z$ as follows:
$$g_x(z_x, z_y, z_c) = (e^{2\pi i/N}z_x, z_y, z_c)$$
$$g_y(z_x, z_y, z_c) = (z_x, e^{2\pi i/N}z_y, z_c)$$
$$g_c(z_x, z_y, z_c) = (z_x, z_y, e^{2\pi i/N}z_c)$$

This action is clearly linear. Furthermore, it clearly satisfies $g \cdot f(w) = f(gw)$. Hence, $f$ is equivariant between the actions on $G$ and $Z$. Furthermore, we can easily see that the invariant subspaces in this case are the first, second, and third coordinate subspaces, and that each subspace is acted on trivially by all but one of the subgroups $G_x$, $G_y$ and $G_c$. Hence the agent's representation is disentangled in our sense.

If we view $\rho$ as a representation over $\mathbb{R}^6$, then the group action will be expressed using block-diagonal matrices of $2 \times 2$ rotation matrices, and the invariant subspaces will be two-dimensional, but the argument will be the same.

\section{Backward compatibility of the new definition}
\label{sec_connections_to_disentanglement}

This section examines whether the new definition fits with the old intuitions on disentangled representations. These intuitions can be characterised along three dimensions: \emph{modularity}, \emph{compactness} and \emph{explicitness}\footnote{These are also sometimes referred to as disentanglement, completeness and informativeness by following the naming notation in \cite{Eastwood_Williams_2018}} \cite{Ridgeway_Mozer_2018}. Different models and different disentanglement metrics have often disagreed on what qualifies as a disentangled representation when considered in terms of these qualities. We are going to show that our definition helps resolve these points of disagreement in a principled manner.

\paragraph{Modularity} Modularity measures whether a single latent dimension encodes no more than a single data generative factor. All previous approaches agree that disentangled representations should have this property. This is also true for our definition, provided that we replace what was previously called the ``data generative factors'' with ``disentangled actions of the symmetry group''. Note that there are examples where our definition may disagree with the past intuitions. For example, consider the case of 3D rotations. Intuitively we might think that the representation could disentangle rotations about the x, y and z axes. However, according to our definition, the representation can only be disentangled with respect to a direct product decomposition of a group. In particular, if the group cannot be factored as a direct product, then the agent's representation cannot be disentangled according to our definition. As discussed in Sec.~\ref{sec_worked_nonlinear_example}, the group of 3D rotations does not decompose into a direct product, and hence a disentangled representation is not possible.

\paragraph{Compactness} Compactness measures whether each data generative factor is encoded by a single latent dimension. This is definitely a point of disagreement. Most of the recent approaches to disentangled representation learning \cite{Higgins_etal_2017, Chen_etal_2018, Kim_Mnih_2018, Burgess_etal_2017, Kumar_etal_2017, Ansari_Soh_2018, Esmaeili_etal_2018}, as well as the majority of the disentanglement metrics \cite{Higgins_etal_2017, Chen_etal_2018, Kim_Mnih_2018, Eastwood_Williams_2018} assume that this property should be true. However, many other approaches \cite{Kulkarni_etal_2015, Denton_2017, Rippel_Adams_2013, Reed_etal_2014, Zhu14, Yang_etal_2015, Goroshin_etal_2015, Cheung15, Whitney_etal_2016, Karaletsos_etal_2016} and a recent disentanglement metric by \cite{Ridgeway_Mozer_2018} consider it perfectly fine for each data generative factor to be encoded by multiple latent dimensions. According to our definition, \emph{each disentangled subspace can be multi-dimensional}. 

\paragraph{Explicitness} Explicitness measures whether the values of \emph{all} of the data generative factors can be decoded from the representation using a \emph{linear} transformation. This is the strongest requirement of the three, since it encompasses two points: that a disentangled representation captures information about \emph{all} the data generative factors, and that this information is \emph{linearly} decodable. Let us consider the two points separately.

In terms of capturing the information about all the data generative factors, there seems to be general agreement among all the approaches and metrics that this is an important aspect of a disentangled representation. Our definition agrees, but with a caveat. Since we define disentangled representations with respect to a particular group decomposition, and groups often have multiple possible decompositions, the same data can induce many different disentangled representations. For example, the maximal decomposition of the symmetry group acting on our grid world example from Sec.~\ref{sec_high_level_overview} is $G = G_x \times G_y \times G_c$. However, the same group can also be decomposed into $G = G_p \times G_c$, where $G_p$ is the group of all position transformations. Hence, a vector representation that consists of two subspaces, one transforming under the colour change action, and the other transforming under the actions of the horizontal and vertical translations will be considered entangled under the former group decomposition, and disentangled under the latter one. Note that this can be beneficial, since different subsequent tasks may benefit from different group decompositions. For example, in order to find a forest one does not need to represent each tree. 

In terms of the linearity aspect of a disentangled representation, it appears to be a point of contention. Most of the models do not optimise for it. However, a few recent disentanglement metrics give reduced disentanglement scores to models that do poorly on this metric \cite{Eastwood_Williams_2018, Kumar_etal_2017, Ridgeway_Mozer_2018}. Linearity is not required for a representation to be disentangled according to our definition, unless we are talking about \emph{linear disentangled representations} discussed in Sec.~\ref{sec_defining_linear_disentangled_representations}.

\section{Conclusions}
\label{sec_discussion}
The aim of this paper was to provide a formal definition of disentangled representations. We started by bringing the insights from modern physics and past work in machine learning to argue that the structure of the world that disentangled representations should capture are the symmetry transformations of the world state. This replaced the ill-defined notion of ``data generative factors'' used in the past. We then used group and representation theory to show how the structure of the symmetry transformations can be reflected in the representation vector space. The resulting insights served as the basis for our definition of disentangled representations. In order to make the connection between symmetry groups and disentangled representations, we had to make a number of simplifying assumptions. In particular, we assumed that the symmetry groups can be decomposed as direct products of subgroups in a natural way and that their interesting decompositions into subgroups were known. We believe that these assumptions can be relaxed in the future, but we leave this to future work. 

We have shown that our definition of disentangled representations fits well with the previously held views on the topic, while also providing a principled way to resolve past disagreements. For example, according to our definition a disentangled subspace can be single- or multi-dimensional, depending on the structure of the symmetry group. Furthermore, we have shown that each disentangled subspace may have many different suitable bases, and the action of the group does not have to be implemented as a linear transformation in the representation vector space, unless a \emph{linear disentangled representation} is required. 

Our framework suggests that the invariant subspaces in the representation vector space are in effect separate representations. So it is as if instead of having a single representation $Z$, the model has several different representations $Z_1$, $Z_2$, ... $Z_n$. They are separate because under the action of the world's dynamics, there is no mixing between them. Hence, it is feasible to assume that the set of representations may be extended in a continual learning scenario, as the model experiences new aspects of the world. In fact, an empirical demonstration of a similar process already exists \cite{Achille_etal_2018}. 

We hope that the insights provided by our proposed definition of disentangled representations will be helpful in speeding up and measuring progress in finding scalable algorithmic solutions to disentangled representation learning.

\subsection*{Acknowledgements}
We thank Chris Burgess, Matt Botvinick, Nick Watters and Pedro Ortega for useful discussions.

\bibliographystyle{abbrv} 
\bibliography{bibliography}

\newpage
\appendix
\section{Appendix}
\label{sec_appendix}

\subsection{Review of group theory}
\label{sec_group_theory}
A \emph{group} $(G, \circ)$ is a set $G$ together with a binary operation $\circ: G \times G \rightarrow G$ satisfying the following axioms:

\begin{enumerate}
    \item Associativity $\forall x,y,z \in G: x \circ (y \circ z) = (x \circ y) \circ z$
    \item Identity $\exists e \in G,  \forall x \in G : e \circ x = x \circ e = x$
    \item Inverse $\forall x \in G, \exists x^{-1} \in G : x \circ x^{-1} = x^{-1} \circ x = e$
\end{enumerate}

Note that the binary operation is not required to be commutative. That is, we need not have $x \circ y = y \circ x, \forall x, y \in G$. A group that is commutative is called Abelian. 

When the binary operation is clear from context, it is customary to omit it, and write $G$ for $(G, \circ)$ and $xy$ for $x \circ y$. It is also customary to refer to the binary operation as \emph{multiplication}, and to write 1 for $e$.

Some important examples of groups are:
\begin{itemize}
    \item The symmetric group $S_n$ of permutations of the numbers $\{1..n\}$. The elements of the group are the bijections from $\{1..n\}$ to itself. The binary operation is function composition.
    \item The general linear group $GL(V)$ of invertible linear transformations of a vector space $V$ under composition. If we have $V = \mathbb{R}^n$ with its usual basis, then this group is also called $GL(n, \mathbb{R})$. From that point of view, its elements are the invertible $n \times n$ matrices with entries in $\mathbb{R}$, and the binary operation is matrix multiplication.
    \item The special orthogonal group $SO(n)$ of distance-preserving transformations of a Euclidean space that also preserve a fixed point. The elements are orthogonal matrices with determinant $1$, and the binary operation is matrix multiplication. $SO(2)$ and $SO(3)$ are widely studied groups of rotations around a point in 2D or a line in 3D respectively.  
\end{itemize}

Given a group $G$, a subgroup of $G$ is a subset $H$ of the elements of $G$ which is closed under multiplication and inverses. That is, $x, y \in H \Rightarrow xy \in H \, \land \, x^{-1} \in H$. For example, $SO(n)$ is a subgroup of $GL(n, \mathbb{R})$.

Given two groups $G$ and $H$, we can construct a new group $G \times H$, called their direct product, as follows:
\begin{enumerate}
    \item The underlying set is the Cartesian product, $G \times H$. That is, the ordered pairs $(g, h)$, where $g \in G$ and $h \in H$
    \item The group operation is defined component-wise:
    $(g_1, h_1) \circ (g_2, h_2) = (g_1 \circ g_2, h_1 \circ h_2)$
\end{enumerate}
The direct product $G \times H$ contains a subgroup $G \times \{e\}$ that may be identified with $G$, and a subgroup $\{e\} \times H$ that may be identified with $H$. Thus we may speak of $G \times H$ as having $G$ and $H$ as subgroups.

Groups often arise as transformations of some space, such as a set, vector space, or topological space. In that case, we speak of the group's \emph{action} on the space. For example, $S_n$ acts on the set $\{1..n\}$ and $GL(V)$ acts on the vector space $V$. Formally, a group action on a set is a function $\cdot : G \times X \rightarrow X$ (hereafter indicated by concatenation) satisfying:
\begin{enumerate}
    \item $e x = x \ \ \forall x \in X$
    \item $(gh)x = g(hx) \ \ \forall g, h \in G, x \in X$
\end{enumerate}
The conditions just say that the action is compatible with the group structure.

When a group acts on a space with additional structure, such as a vector space or a topological space, then we also require that the action preserves that structure. In the case of a vector space, the requirement would be that the action is linear:
\begin{enumerate}
    \item $g(x + y) = g x + g y \ \ \forall g \in G, x, y \in V$
    \item $g(\lambda x) = \lambda (g x) \ \ \forall g \in G, \lambda \in \mathbb{R}, x \in V$
\end{enumerate}
In the case of a topological space, the requirement would be that the action is continuous, and so on.

When a group $G$ acts on a space $X$, then we say informally that the elements of $G$ are \emph{symmetries} of $X$.

\subsection{Review of group representation theory}
\label{sec_representation_theory}
The axiomatic definition of groups outlined in Appendix~\ref{sec_group_theory} allows for any abstract structure to be a group. In order to understand these abstract groups, mathematicians have found it helpful to relate them to linear algebra. \emph{Group representation theory} studies abstract groups by \emph{representing} their elements as linear transformations of vector spaces. Concretely, a \emph{group representation} is a function $\rho: G \rightarrow GL(V)$ satisfying:
\begin{itemize}
    \item $\rho(gh) = \rho(g)\rho(h)$
    \item $\rho(e) = 1_V$ 
\end{itemize}

Note that in these equations there are two groups in play, $G$ and $GL(V)$. On the left hand side of each equation, the multiplication or inverse takes place in $G$. On the right hand side it takes place in $GL(V)$. What the equations say is that the function $\rho$ preserves the group structure.

Relative to a given basis for $V = \mathbb{R}^n$, group elements are represented as invertible $n \times n$ matrices in $GL(n, \mathbb{R})$. If the representation $\rho$ is injective, then this allows us to think of $G$ as being a subgroup of $GL(n, \mathbb{R})$. However, $\rho$ need not be injective in general. For example, every group has a trivial representation which sends every element to the identity.

Note that when $\rho$ is clear from context, we will sometimes omit it and refer to $V$ as being the representation.

Another way to think about a group representation is that it is a group action $\cdot : G \times V \rightarrow V$ on a vector space that acts linearly. Given a representation $\rho: G \rightarrow GL(V)$, the action is defined as $\cdot: G \times V \rightarrow V$, $\cdot: (g, v) \mapsto \rho(g)(v)$. We shall switch freely between these two points of view as convenient.

Given two representations $\rho_1: G \rightarrow GL(V)$, $\rho_2: G \rightarrow GL(W)$ of the same group, we can construct a new representation, their direct sum. This is a representation $\rho_1 \oplus \rho_2: G \rightarrow GL(V \oplus W)$, of linear transformations acting on the direct sum $V \oplus W$ of the two vector spaces. It is defined in the obvious way:
$$(\rho_1 \oplus \rho_2)(g)(v, w) = (\rho_1(g)(v), \rho_2(g)(w))$$ 
In matrix terms, if $\rho_1$ represents $g \in G$ as the matrix $A$ and $\rho_2$ represents it as $B$, then $\rho_1 \oplus \rho_2$ represents $g$ as the block-diagonal matrix
\[
(\rho_1 \oplus \rho_2)(g) = \begin{bmatrix}
A & 0 \\ 
0 & B
\end{bmatrix}
\]

Given a representation $\rho: G \rightarrow GL(V)$, a subrepresentation is any vector subspace $W \leq V$ that is invariant under the action of $G$. That is, $\rho(g)(w) \in W \ \ \forall g \in G, w \in W$. Under certain conditions (for example, if $G$ is finite), then there exists another invariant subspace $W^\perp$ such that $V = W \oplus W^\perp$.

Every representation $V$ has the trivial subspace $\{0\}$ and V itself as subrepresentations. If V has other subrepresentations then it is said to be \emph{reducible}. Otherwise it is irreducible. If $V$ is reducible, then we may recursively decompose its invariant subspaces, until we get a direct sum $V = W_1 \oplus W_2 \oplus ... \oplus W_n$, where each of the $W_i$ is an irreducible representation. By suitable choice of basis, the matrices $\rho(g)$ will then be block-diagonal, as above.

It may appear that a representation could have different decompositions into irreducibles, depending for example on which invariant subspace we chose first. However it turns out that (under certain conditions) this is not true: for a given representation $V$, it doesn't matter how we decompose it, we will always end up with the same set of irreducible representations $W_i$, up to isomorphisms, order and change of basis.

There is another construction that is important for our discussion. We can construct a tensor product representation $\rho_1 \otimes \rho_2: G \rightarrow GL(V \otimes W)$, acting on the tensor product $V \otimes W$ of the two vector spaces.

If $V$ has a basis $\{e_i | i \in \{1..m\}\}$ and $W$ has a basis $\{f_j | j \in \{1..n\}\}$, then the tensor product $V \otimes W$ is a vector space with a basis $\{e_i \otimes f_j | i \in \{1..m\}, j \in \{1..n\}\}$. It is an $m \times n$-dimensional vector space. This gives rise to a tensor product representation via:
$$(\rho_1 \otimes \rho_2)(g)(v \otimes w) = \rho_1(g)(v) \otimes \rho_2(g)(w)$$ 

If we consider the basis $\{e_i \otimes f_j\}$ as being ordered lexicographically, then in matrix terms, if $g \in G$ is represented by $\rho_1$ as $A$ and by $\rho_2$ as $B$, then in the tensor product representation it is represented as 
\[
(\rho_1 \otimes \rho_2)(g) = \begin{bmatrix}
a_{11} B & a_{12} B & ... & a_{1m} B \\ 
a_{21} B & ... & ... & a_{2m} B \\ 
... & ... & ... & ... \\ 
a_{m1}  B & a_{m2}  B & ... & a_{mm}  B 
\end{bmatrix}
\]

Note that if $V$ is one-dimensional, then $V \otimes W$ can be identified with $W$. If moreover $\rho_1$ is the trivial representation that sends everything to the identity, then every $g \in G$ will be represented by the matrix $A = [1]$, and $\rho_1 \otimes \rho_2$ will have the same matrix as $\rho_2$. Similarly, if $W$ is one-dimensional and $\rho_2$ is the trivial representation, then $\rho_1 \otimes \rho_2$ will be the same as $\rho_1$.

Finally, we will need the following result: If our group is a direct product $G = G_1 \times G_2$, then the irreducible representations of $G$ are precisely the representations $\rho_1 \otimes \rho_2$, for $\rho_1, \rho_2$ irreducible representations of $G_1, G_2$ respectively (see Lemma~22.6 in \cite{Teleman_2005}).

\end{document}